\documentclass[10pt,twocolumn,letterpaper]{article}

\usepackage[pagenumbers]{cvpr} % To force page numbers, e.g. for an arXiv version

\usepackage{times}
\usepackage{epsfig}
\usepackage{graphicx}
\usepackage{amsmath}
\usepackage{amssymb}
\usepackage{booktabs}
\usepackage{color}
\usepackage{caption}
\usepackage{subcaption}
\usepackage{makecell}
\usepackage{multirow}

\usepackage{bbding}
\usepackage{url}

\usepackage{float}
\usepackage{stfloats}
\usepackage{balance}

\usepackage{tcolorbox}
\usepackage{xcolor}
\usepackage{circledsteps}

\definecolor{cvprblue}{rgb}{0.21,0.49,0.74}
\definecolor{salmon}{RGB}{255,103,125}
\usepackage[pagebackref,breaklinks,colorlinks,allcolors=cvprblue]{hyperref}

\def\paperID{2004} % *** Enter the Paper ID here
\def\confName{CVPR}
\def\confYear{2026}

\title{\textbf{\texttt{Arcadia}}: Toward a Full-Lifecycle Framework for Embodied Lifelong Learning}

\author{
Minghe Gao\textsuperscript{1,2}\footnotemark[1]\thanks{Equal Contribution} \ \thanks{Work done when interning at Unitree}, 
Juncheng Li\textsuperscript{1}\thanks{Corresponding author.}, 
Yuze Lin\textsuperscript{1}\footnotemark[1], 
Xuqi Liu\textsuperscript{1}\footnotemark[1], 
Jiaming Ji\textsuperscript{3}\footnotemark[1]\\
Xiaoran Pan\textsuperscript{1}, 
Zihan Xu\textsuperscript{1}, 
Xian Li\textsuperscript{1}, 
Mingjie Li\textsuperscript{2}, 
Wei Ji\textsuperscript{4},
Rong Wei\textsuperscript{5},
Rui Tang\textsuperscript{5}, \\
Qizhou Wang\textsuperscript{2}, 
Kai Shen\textsuperscript{6}, 
Jun Xiao\textsuperscript{1}, 
Qi Wu\textsuperscript{7}, 
Siliang Tang\textsuperscript{1}, 
Yueting Zhuang\textsuperscript{1}
\\
{\small \textsuperscript{1}Zhejiang University \quad \textsuperscript{2}Unitree Tech \quad \textsuperscript{3}Peking University \quad \textsuperscript{4}Nanjing University} \\ 
{\small \textsuperscript{5}Manycore Tech \quad \textsuperscript{6}Bytedance Seed \quad \textsuperscript{7}University of Adelaide} 
\\
{\small \url{https://github.com/Embodied-Arcadia/EmbodiedKit/}}
}

\begin{document}
\maketitle

\begin{figure*}[t]
    \centering
    \vspace{-4mm}
    \includegraphics[width=\linewidth]{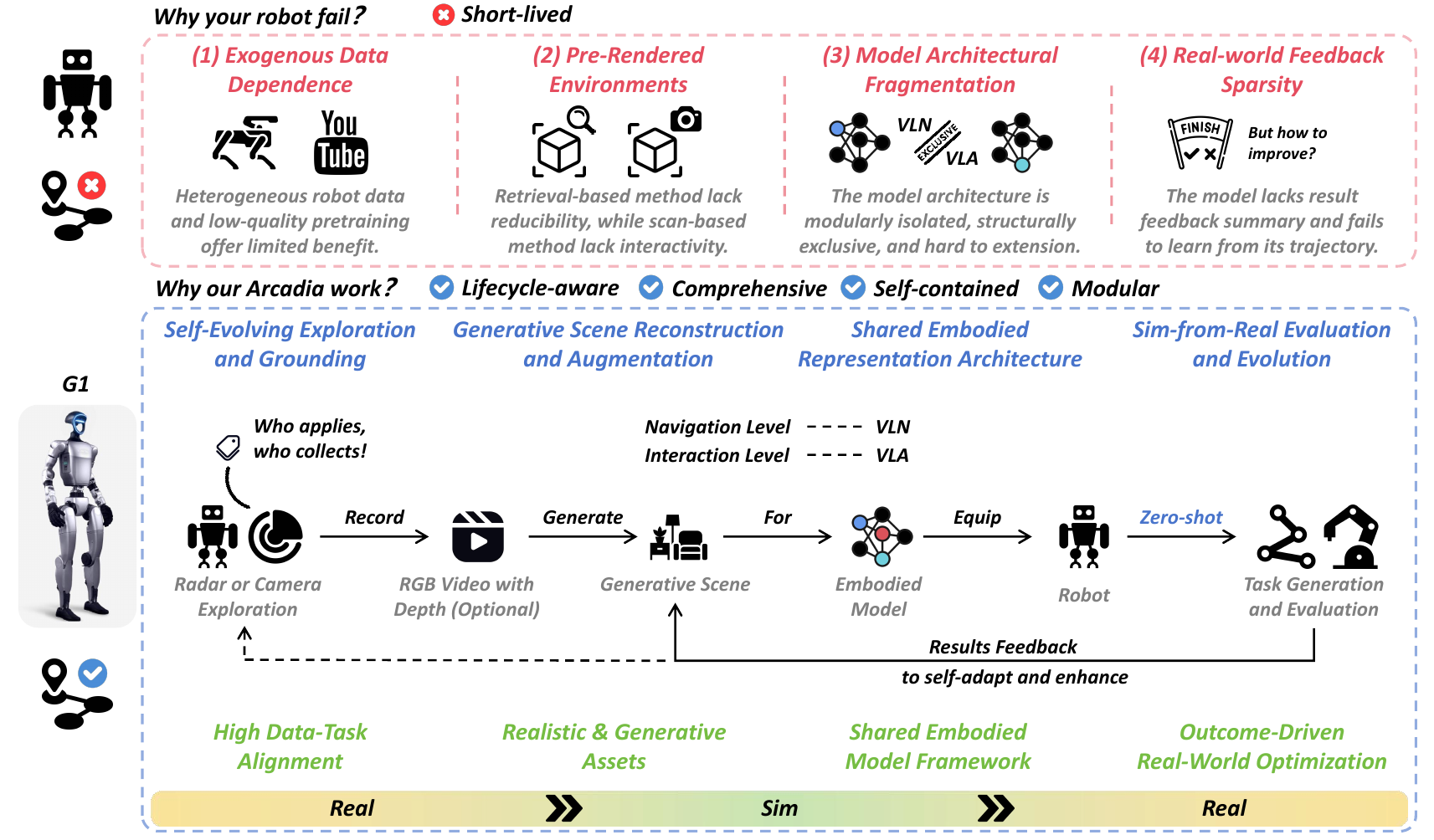}
    \vspace{-6mm}
    \caption{\textbf{\texttt{Arcadia}} provides an overview of a full real-to-sim-to-real lifecycle for embodied learning, illustrating how the framework closes the loop between real-world experience, simulation, and redeployment while addressing four core limitations in contemporary embodied AI: exogenous data dependence, static pre-rendered environments, fragmented model architectures, and sparse real-world feedback.}
    \vspace{-2mm}
    \label{fig1}
\end{figure*}

\begin{abstract}
We contend that embodied learning is fundamentally a lifecycle problem rather than a single-stage optimization. Systems that optimize only one link (data collection, simulation, learning, or deployment) rarely sustain improvement or generalize beyond narrow settings. We introduce \textbf{\texttt{Arcadia}}, a closed-loop framework that operationalizes embodied lifelong learning by tightly coupling four stages: (1) Self-evolving exploration and grounding for autonomous data acquisition in physical environments, (2) Generative scene reconstruction and augmentation for realistic and extensible scene creation, (3) a Shared embodied representation architecture that unifies navigation and manipulation within a single multimodal backbone, and (4) Sim-from-real evaluation and evolution that closes the feedback loop through simulation-based adaptation. This coupling is non-decomposable: removing any stage breaks the improvement loop and reverts to one-shot training. \textbf{\texttt{Arcadia}} delivers consistent gains on navigation and manipulation benchmarks and transfers robustly to physical robots, indicating that a tightly coupled lifecycle: continuous real-world data acquisition, generative simulation update, and shared-representation learning, supports lifelong improvement and end-to-end generalization. We release standardized interfaces enabling reproducible evaluation and cross-model comparison in reusable environments, positioning \textbf{\texttt{Arcadia}} as a scalable foundation for general-purpose embodied agents.
\end{abstract}

\vspace{-3mm}
\section{Introduction}

Humans learn across an entire lifespan, accumulating and reorganizing experience to enable broad generalization and skill transfer beyond the training context~\cite{baltes1987theoretical,baltes1997incomplete,kolb2014experiential,hebb2005organization,lindenberger2019brain}. 
Cognitive and embodied theories further argue that competence is shaped by continuous perception–action coupling over time, not by short-lived skills in isolation~\cite{parisi2019continual,chen2018lifelong,wilson2013embodied,sullivan2018learning}. This perspective inspires implications for robotics: optimizing only one link of the chain (\textit{e.g.}, training in static simulators or deploying without feedback) rarely sustains improvement or robustness~\cite{pfeifer2006body,xiao2025robot}. Recent studies show that coordinating multiple stages of the embodied pipeline (real-world data collection, generative simulation, unified representation learning, and deployment-time feedback) yields stronger generalization and data efficiency than siloed designs~\cite{mon2025embodied,zitkovich2023rt,o2024open,team2024octo,bharadhwaj2024roboagent,bousmalis2023robocat}. We adopt this view and refer it as \textbf{embodied lifelong learning}: designing and evaluating embodied intelligence as closed loops that continually couple real experience with updates to representations, policies, and the simulators in which they are trained.

Recent efforts have begun to integrate previously isolated stages.
GRUtopia~\cite{wang2024grutopia} unifies diverse scenarios, agents, and benchmarks within a single simulation system, while NaVILA~\cite{cheng2024navila} connects high-level language instructions to low-level motor control and validates them on physical robots to mitigate the sim-to-real gap. These works expand the coverage of the pipeline, yet they still fall short of a full lifecycle. GRUtopia primarily extends the simulation side of the pipeline, while NaVILA primarily extends the execution span toward real robots. Both broaden coverage but stop short of closing the lifecycle: they do not establish a persistent, editable path from deployment experience back into simulation assets, nor a persistent mechanism for routing in-deployment signals to supervise subsequent training. As a result, they connect individual edges without maintaining the full real-to-sim-to-real loop across stages.

On the real-to-sim side, two key limitations dominate:
\textbf{\textcolor{salmon}{(1) Exogenous Data Dependence.}}
From a lifecycle perspective, training data are most effective when sourced from, and remain grounded in, deployment contexts. Reliance on external, off-distribution sources weakens this coupling: YouTube videos are misaligned with embodied control objectives, while data from quadruped robots reflects a distinct morphology and viewpoint relative to a humanoid. Such exogenous and noisy data introduce distribution mismatches and are costly to curate or modify in simulation, let alone to adapt based on feedback. It hinders execution signals from driving asset or policy updates and stalling lifelong improvement. This limitation is evident in NaVILA~\cite{cheng2024navila}, which compensates for data scarcity using mismatched external corpora,  yielding limited gains.
\textbf{\textcolor{salmon}{(2) Pre-Rendered Environments.}} 
Under a lifecycle view, simulators should ingest real observations and remain editable so that deployment experience can be incorporated into new scenes. Pre-rendered or static environments (e.g., Matterport3D~\cite{chang2017matterport3d} and Habitat v1~\cite{savva2019habitat}) with limited physical properties, restrict controllability and prevent the faithful insertion of deployment-time variations, slowing or even blocking the data–asset–policy loop. In practice, the common workflow, which trains on static sets and then validates within Isaac Sim~\cite{nvidia_isaac_sim}/Lab~\cite{nvidia_isaac_lab}, adds engineering overhead without restoring lifecycle coupling. Even large, high-fidelity suites such as GRUtopia~\cite{wang2024grutopia} rely on finite asset libraries and retrieval-based scene variations, limiting generative adaptation and hindering sustained improvement.

On the sim-to-real side, two further limitations appear: 
\textbf{\textcolor{salmon}{(3) Model Architectural Fragmentation.}}
A lifecycle system benefits from a shared embodied representation so that data, supervision, and deployment feedback can propagate across task families and embodiments~\cite{jaafar2024lanmp,shridhar2020alfred,padmakumar2022teach}. In practice, pipelines often build separate task-specific stacks with incompatible assumptions (e.g., navigation agents modeled as oriented bounding boxes in indoor simulators~\cite{anderson2018vision,ku2020room} and tabletop manipulation under fixed camera and end-effector control~\cite{shridhar2022cliport,james2020rlbench}), which block cross-task credit assignment and prevent coherent policy updates from real deployments. The result is siloed improvement for long-horizon, multi-skill objectives, which is the opposite of lifecycle coupling.
\textbf{\textcolor{salmon}{(4) Real-world Feedback Sparsity.}} 
In a lifecycle system, deployment should provide dense, structured signals that update both simulator assets and policies. Yet many pipelines treat deployment as one-shot, capturing at most coarse success/failure labels~\cite{ross2011reduction,meng2025preserving}. This sparsity stalls the cycle: long-horizon errors cannot be localized, partial progress and environmental shifts are not fed back, and execution confidence cannot guide refinement. Effective lifecycle coupling requires routing multi-level feedback (task progress, scene dynamics, hardware constraints) back into asset updates and policy learning, rather than treating deployment as the endpoint.

We summarize these four limitations in Figure~\ref{fig1}. Taken together, they point not only to isolated algorithmic gaps but to a breakdown of lifecycle coupling: from data acquisition through simulation, representation, and deployment-time supervision. A platform oriented to embodied lifecycle learning should therefore (i) maintain strong alignment between collected experience and target tasks, (ii) translate real observations into editable, generative simulation assets, (iii) learn with a shared, extensible embodied representation across tasks, and (iv) route outcome-driven, in-deployment feedback back into both assets and policies.

To address these challenges, we present \textbf{\texttt{Arcadia}, a full-lifecycle framework for embodied lifelong learning in a closed loop}. Specifically, it comprises four components:  
\textbf{\textcolor{cvprblue}{(1) Self-Evolving Exploration and Grounding}}, in which robots autonomously collect task-aligned data within their physical environments;  
\textbf{\textcolor{cvprblue}{(2) Generative Asset Reconstruction and Augmentation}}, which transforms real-world sensory data into realistic and extensible simulation assets;  
\textbf{\textcolor{cvprblue}{(3) Shared Embodied Representation Architecture}}, enabling visual-language-navigation (VLN) and visual-language-action (VLA) model to share a common multimodal backbone; and  
\textbf{\textcolor{cvprblue}{(4) Sim-from-Real Evaluation and Evolution}}, which integrates real-world feedback into simulation as adaptive supervision to continuously refine embodied behavior.

Crucially, the four components function both independently and as a coupled closed-loop system. Each addresses a distinct bottleneck in the embodied intelligence lifecycle while collectively driving continual self-improvement. This design yields substantial gains across both simulated and physical domains. In simulator evaluations, \textbf{\texttt{Arcadia}} achieves an average improvement of 7.07\% on navigation and 11.08\% on manipulation tasks compared with other baseline methods. Furthermore, in real-world experiments conducted on a Unitree G1 robot equipped with a Dex-3 manipulator, \textbf{\texttt{Arcadia}} successfully completed 46\% of navigation and 27\% of manipulation trials, substantially outperforming NaVILA (13\%) and OpenVLA (9\%). These improvements are particularly evident in multi-destination navigation and multi-object manipulation scenarios. Collectively, these results highlight \textbf{\texttt{Arcadia}}’s promising generalization in real-world embodied operation. We also open-source standardized interfaces for benchmarking and cross-model evaluation within reusable environments.

\noindent\textbf{In summary, our contributions are as follows:}
\begin{itemize}
    \item We propose \textbf{\texttt{Arcadia}}, a full-lifecycle embodied intelligence framework that unifies data, simulation, learning, and feedback into a closed self-improving loop.
    \item We develop a real-to-sim generative pipeline that reconstructs real assets for scalable, high-fidelity simulation.
    \item We introduce a novel \textit{Sim-from-Real} feedback mechanism that transfers structured real-world signals back into simulation for continual policy and asset refinement, effectively closing the real-to-sim-to-real loop.
\end{itemize}

\section{Related Works}

\subsection{Embodied Lifelong Learning}
Recent advances in embodied intelligence extend lifelong learning beyond static datasets toward continuously evolving perception–action loops. Rather than optimizing fixed policies in isolated tasks, embodied lifelong learning seeks persistent cycles that link data acquisition, model adaptation, and deployment feedback. This perspective has inspired systems combining autonomous exploration, self-supervised refinement, and multi-task generalization across diverse embodiments~\cite{parisi2019continual,de2021continual,khetarpal2022towards,jaafar2024lanmp,shridhar2020alfred,padmakumar2022teach}. Several frameworks have explored continual adaptation within real or simulated settings such as Octo, RoboCat, and RoboAgent which enable robots to expand skills across domains through ongoing experience~\cite{team2024octo,bousmalis2023robocat,bharadhwaj2024roboagent}. These efforts mark a shift from episodic to adaptive embodied learning.

However, most pipelines remain fragmented: data, training, and deployment are loosely coupled, and deployment feedback rarely supervises further learning. \textbf{\texttt{Arcadia}} adopts a closed-lifecycle formulation, in which real experience continuously informs simulation, representation, and policy updates, enabling sustained self-improvement across the full perception–action loop.

\subsection{Bridging Simulation and Reality}
Because large-scale real-world experimentation is costly and risky, embodied learning often maps physical environments into simulation, where agents can explore efficiently. This results in a reality gap: the mismatch between simulated and physical domains~\cite{jakobi1995noise,tobin2017domain,peng2018sim}. Domain randomization over textures, lighting, and dynamics improves robustness and has powered scalable systems like OpenAI’s Dactyl~\cite{andrychowicz2020learning}, yet handcrafted simulators capture only limited real-world complexity, such as deformable objects, sensor noise, or dynamic scenes, leaving trained policies brittle after deployment.

Recent work has begun to narrow this gap through generative reconstruction and adaptive simulation frameworks~\cite{wang2024grutopia,cheng2024navila}, which translate real-world observations into editable, physics-consistent virtual assets. \textbf{\texttt{Arcadia}} extends this direction toward a bidirectional, closed-loop paradigm: real-world data regenerates simulation assets and supervision, while feedback from simulation refines shared representations and policies before redeployment. This continual real-to-sim-to-real cycle transforms simulation from a static proxy into an active driver of embodied adaptation.

\section{Method}

\begin{figure*}[t]
    \centering
    \vspace{-4mm}
    \includegraphics[width=1\linewidth]{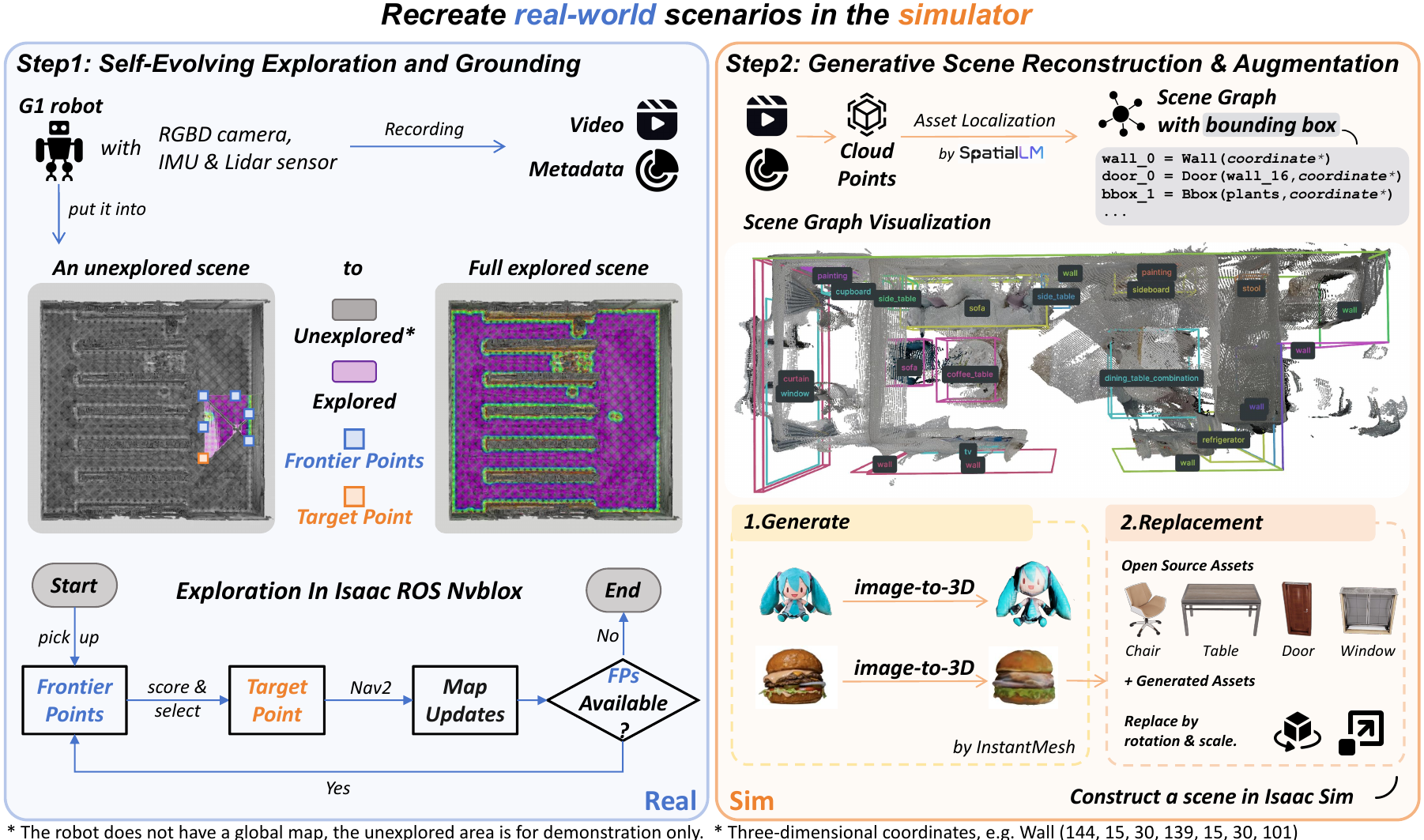}
    \vspace{-6mm}
    \caption{Overview of \textbf{\texttt{Arcadia}}’s real-to-sim pipeline: robots autonomously explore real environments to collect multimodal data (Step~1), which are then reconstructed and augmented into editable 3D scenes for simulation (Step~2).}
    \vspace{-3mm}
    \label{fig2}
\end{figure*}

\begin{figure*}[t]
    \centering
    \vspace{-4mm}
    \includegraphics[width=1\linewidth]{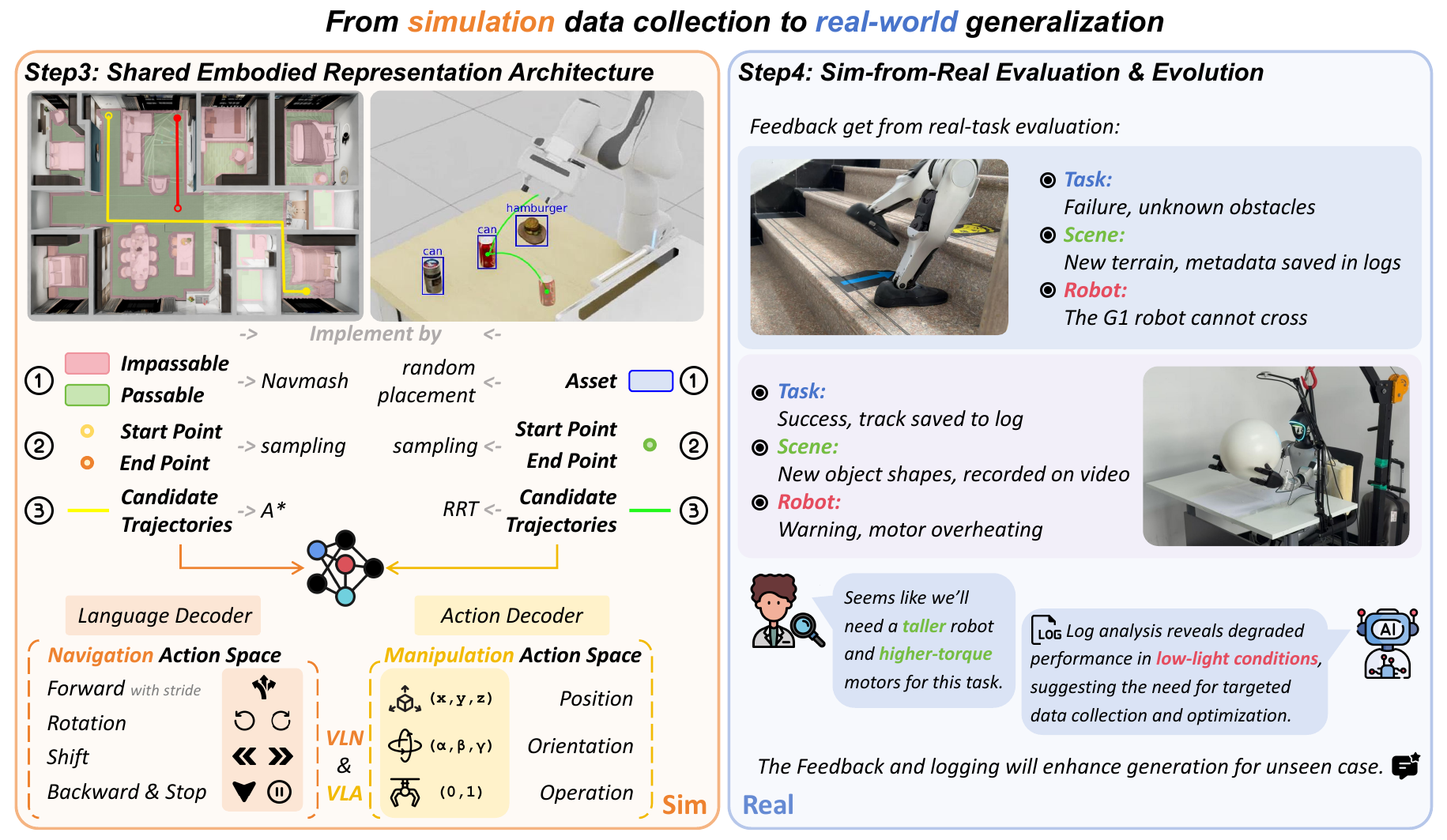}
    \vspace{-6mm}
    \caption{Overview of \textbf{\texttt{Arcadia}}’s sim-to-real pipeline: navigation and manipulation trajectories are collected in simulation (Step 3) using A* and RRT planning within a shared embodied architecture, then real-world feedback is integrated for continual refinement (Step 4).}
    \label{fig3}
    \vspace{-3mm}
\end{figure*}

Given a natural-language instruction such as “bring me the cup from the table”, \textbf{\texttt{Arcadia}} executes a complete real-to-sim-to-real cycle (Figure~\ref{fig2} and~\ref{fig3}). The process starts with autonomous exploration and data collection in real environments (Sec.~\ref{sec3.1}), followed by generative reconstruction of realistic, editable simulation assets (Sec.~\ref{sec3.2}). The simulator then trains unified navigation and interaction policies under a shared embodied model (Sec.~\ref{sec3.3}). Finally, real-world deployment provides feedback that is incorporated back into simulation to refine both assets and policies (Sec.~\ref{sec3.4}). This closed loop enables continual, self-improving embodied learning across the robot’s lifecycle.

\subsection{Self-Evolving Exploration and Grounding}
\label{sec3.1}

Conventional embodied datasets, collected from exogenous sources or passive mapping, often exhibit weak task alignment and domain gaps between training and deployment. To bridge this, \textbf{\texttt{Arcadia}} autonomously acquires task-relevant data in the same physical environments used for deployment, ensuring that perception and control models learn under realistic conditions.

Built on Isaac ROS~\cite{isaac_ros} and Nvblox~\cite{isaac_ros_nvblox} for SLAM and 3D reconstruction, our method employs a frontier-based exploration policy that maximizes information gain. Frontier points, the boundaries between explored and unexplored regions, are scored by expected entropy reduction, and the robot visits the highest-scoring ones using low-level motion control APIs. The map and frontier sets are continuously updated, producing adaptive trajectories that balance coverage, efficiency, and semantic relevance. Compared with grid or scripted exploration, this self-evolving policy emphasizes regions critical to downstream tasks, improving both sample efficiency and task-grounded coverage.

After exploration, \textbf{\texttt{Arcadia}} outputs synchronized multimodal data (RGB-D, LiDAR, IMU, odometry, poses). Retaining complete observation histories instead of discarding intermediate frames yields dense, temporally grounded supervision for reconstruction and policy learning, maintaining lifecycle consistency and reducing the real-to-sim gap.

\subsection{Generative Scene Reconstruction}
\label{sec3.2}

Traditional simulators rely on static scans or retrieval-based scene assembly, which require manual curation and offer limited domain coverage. To overcome these constraints, \textbf{\texttt{Arcadia}} employs a generative reconstruction pipeline that converts real environments into editable, task-aligned simulation assets, combining the realism of physical data with the scalability of synthesis.

From the multimodal inputs collected in Section~\ref{sec3.1}, videos and point clouds are parsed into a structured 3D scene graph $G=(V,E)$, with objects/architectural elements as nodes and spatial relations as edges, implemented by a scene parsing module (\textit{e.g.}, SpatialLM~\cite{mao2025spatiallm}). Rather than retrieving meshes from databases, \textbf{\texttt{Arcadia}} synthesizes assets directly from multi-view observations using a Gaussian-splat-based reconstructor~\cite{xu2024instantmesh}, producing simulator-compatible USD objects with consistent geometry, texture, and semantics. This enables broad domain expansion without manual intervention, reduces asset bias, and preserves task semantics observed in the real world.

All assets are imported into Isaac Sim~\cite{nvidia_isaac_sim} via an automated management interface, yielding high-fidelity, extensible environments. By replacing manual retrieval with generative synthesis, \textbf{\texttt{Arcadia}} delivers realistic yet diverse simulations that support scalable, lifelong embodied learning while maintaining lifecycle consistency.

\subsection{Shared Embodied Representation Architecture}
\label{sec3.3}

Most embodied systems train different locomotion and manipulation tasks as separate models with independent perception stacks, action spaces, and objectives. This separation imposes a hard boundary between spatial reasoning and fine-grained interaction, resulting in two issues: (i) no shared grounding across language, vision, and action, requiring each model to relearn objects, relations, and affordances; and (ii) brittle cross-stage coupling, where navigation can reach the vicinity of a goal but exposes no structured state for manipulation, while manipulation assumes an arm-centric viewpoint. This fragmentation increases data requirements, limits transfer, and degrades performance on long-horizon language-conditioned instructions.

\textbf{\texttt{Arcadia}} instead employs a unified multimodal backbone jointly trained across navigation and manipulation tasks, with lightweight task-specific decoders for action generation. Supervision is generated in simulation: for navigation, start–goal pairs are sampled and A* produces collision-free paths expressed in a 7-primitive discrete control space, which generalizes across robot morphologies; for manipulation, RRT generates physically feasible trajectories. All trajectories are language-conditioned at the input, formatted as in VLN-CE~\cite{krantz2020beyond} and BridgeData V2~\cite{walke2023bridgedata}, and passed through the shared perception/state encoder into their respective decoders.

Joint training over both tasks induces a shared embodied state representation, global layout, reachable goals, and approach strategies from navigation are encoded in the same latent space as local affordances and contact behaviors from manipulation. This shared grounding reduces modality drift and facilitates representation transfer between tasks. While each task is evaluated independently under its corresponding embodiment, this unified backbone establishes a common semantic and spatial understanding, enabling coherent reasoning over long-horizon instructions.

\subsection{Sim-from-Real Evaluation and Evolution}
\label{sec3.4}

Most embodied pipelines terminate after inference and treat deployment as purely evaluative, discarding dense execution traces. \textbf{\texttt{Arcadia}} instead treats deployment as an additional supervision stage: real-world rollouts are logged, decomposed into structured feedback signals, and reintegrated into simulation to update both policy and environment. This closes the loop between real and simulated data and enables continual adaptation rather than static training.

We extract feedback along three channels. \textbf{Task feedback.} Each task is decomposed into step-level actions, and feedback at time $t$ is defined as
\[
F_t^{T} = \lambda_1 R_t + \lambda_2 \| s_{t+1} - s_t \| + \lambda_3 \mathcal{L}_{\text{conf}}(o_t, \hat{o}_t) + \lambda_4 \mathcal{L}_{\text{goal}}(s_t, s_g),
\]
where $R_t$ is the scalar reward, $\| s_{t+1} - s_t \|$ measures state transition magnitude, $\mathcal{L}_{\text{conf}}$ measures perceptual consistency between predicted and observed observations, and $\mathcal{L}_{\text{goal}}$ measures distance to the goal state; $\lambda_i$ weight these terms. This converts raw trajectories into a supervisory signal that jointly encodes reward, dynamics, perception, and goal alignment, allowing both global rollout scoring and localized error attribution.

\begin{table*}
\centering 
\vspace{-3mm}
\caption{Comparison of various methods on different benchmarks.}
\vspace{-3mm}
\label{tab1}
\setlength{\tabcolsep}{9pt}
\resizebox{\textwidth}{!}{
\begin{tabular}{l | c c c c c c c c c c c c c c c c} 
\midrule
 & \multicolumn{4}{c}{VLN-CE-Isaac} & \multicolumn{4}{c}{R2R Val-Unseen} & \multicolumn{4}{c}{RxR Val-Unseen} & ScanQA \\
  \cmidrule(lr){2-5}  \cmidrule(lr){6-9} \cmidrule(lr){10-13} \cmidrule(lr){14-14}
  & NE↓ & OS↑ & SR↑ & SPL↑  & NE↓ & OS↑ & SR↑ & SPL↑ & NE↓ & OS↑ & SR↑ & SPL↑ & Meteor↑\\
\midrule
Tuning & 6.12 & 50.7 & 44.9 & 38.5 & 5.44 & 56.0 & 47.1 & 41.3 & 7.12 & 45.2 & 40.9 & 53.7 & 13.4\\
NaVILA & 5.94 & 53.2 & 45.1 & 40.1 & 5.30 & 61.0 & 51.6 & 47.5 & 6.78 & 47.8 & 42.6 & 57.8 & 16.3\\
\textbf{\texttt{Arcadia}} w/o feedback & 5.51 & 56.9 & 48.7 & 43.6 & 5.18 & 62.0 & 54.2 & 49.4 & 6.71 & 49.4 & 44.5 & 58.3 & 19.0 \\
\textbf{\texttt{Arcadia}} w/ feedback & 5.32 & 58.6 & 50.1 & 45.0 & 5.03 & 64.5 & 55.9 & 49.8 & 6.50 & 50.3 & 45.7 & 60.1 & 19.1\\
\midrule
\end{tabular}
}
\end{table*}

\begin{figure*}
    \centering
    \includegraphics[width=1\linewidth]{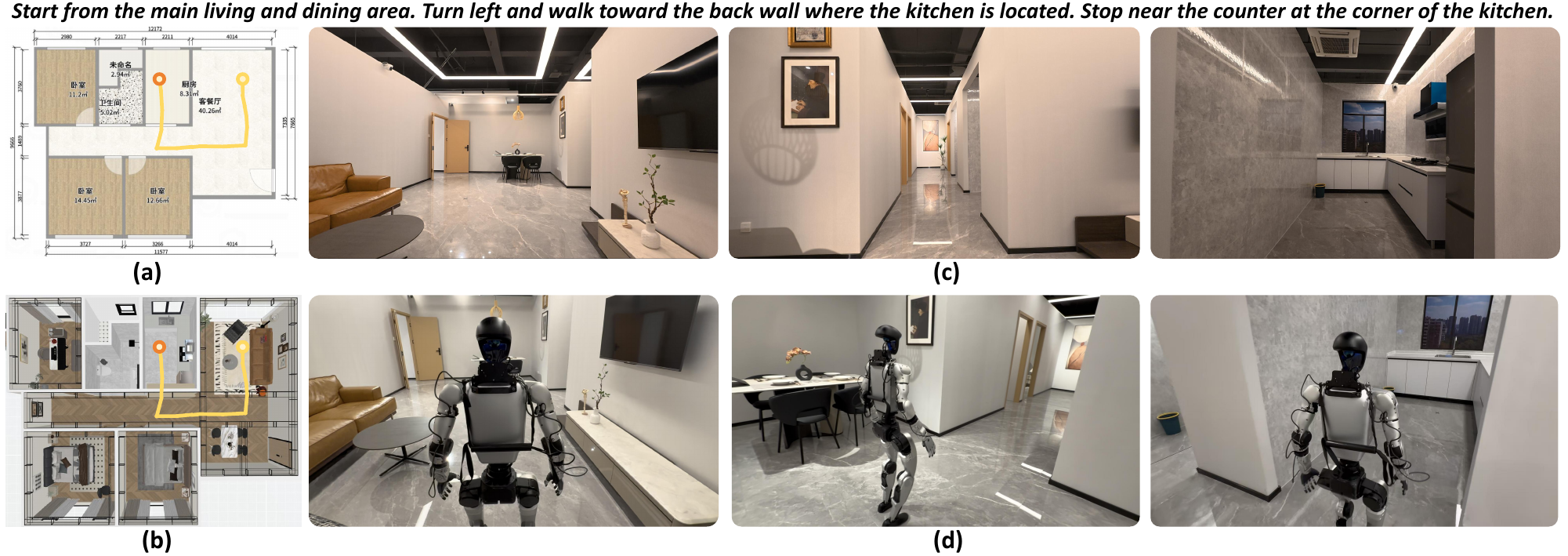}
    \vspace{-7mm}
    \caption{Navigation route from the living and dining area to the kitchen by Unitree G1 robot. (a) Floor plan showing the designated path. (b) 3D rendering of the environment with the same route. (c) First-person visual sequence along the path. (d) Third-person view.}
    \label{fig4}
    \vspace{-3mm}
\end{figure*}

\textbf{Scene feedback.} Multimodal sensory streams (RGB, depth, LiDAR, IMU) are used to characterize environment dynamics and perception quality. Failures such as degraded mapping in low light or the appearance of previously unseen objects are logged and used to update the simulator: new assets are instantiated or perturbations are injected, so that future simulation reflects deployment conditions rather than a fixed, pre-rendered scene.

\textbf{Robot feedback.} Hardware telemetry (joint states, actuator load, communication stability) is monitored during execution. Constraint violations (\textit{e.g.}, a robot exceeding allowable step height or a manipulator exceeding payload limits) are recorded as a robot-level signal $F^{R}$, which is used both for safety gating and for adapting motion policies to platform limits.

These feedback channels are fed back into simulation to update assets, dynamics, and supervision targets, producing a bidirectional real-to-sim-to-real loop. In \textbf{\texttt{Arcadia}}, policy learning, environment generation, and evaluation are no longer disjoint stages but part of the same iterative update process, reducing the sim-to-real gap at training time rather than compensating for it at deployment.

\section{Experiments}

We evaluate \textbf{\texttt{Arcadia}} across both simulation and real-world settings to assess its effectiveness throughout the embodied intelligence lifecycle. Our experiments aim to answer four key questions:  
\textbf{(Q1)} Does \textbf{\texttt{Arcadia}} improve performance on VLN tasks?  
\textbf{(Q2)} Does it enhance VLA tasks involving object manipulation?  
\textbf{(Q3)} How well does it transfer to physical robots in real-world environments?  
\textbf{(Q4)} Which components contribute most to its performance, as shown by ablation studies? 
We address these questions through standardized benchmarks, controlled robotic deployments, and systematic analyses.

\subsection{VLN Task Performance}
\label{sec4.1}

\textbf{Settings.} We evaluate the impact of \textbf{\texttt{Arcadia}} on VLN performance under the following four training configurations:  
\begin{itemize}
\item \textit{Tuning.} Fine-tuning the backbone model solely on VLN-CE~\cite{krantz2020beyond} data in a single-stage setup.  
\item \textit{NaVILA}~\cite{cheng2024navila}. Extends \textit{Tuning} by adding an auxiliary QA fine-tuning stage to improve language grounding.  
\item \textit{\textbf{\texttt{Arcadia}} w/o feedback.} Uses the same multi-stage setup as \textit{NaVILA} but replaces first-stage trajectories with task-aligned data autonomously collected by \textbf{\texttt{Arcadia}}.  
\item \textit{\textbf{\texttt{Arcadia}} w/ feedback.} The full \textbf{\texttt{Arcadia}} framework, where real-world feedback further refines navigation data and policies, completing the real-to-sim-to-real loop.  
\end{itemize}

\begin{figure*}
    \centering
    \vspace{-5mm}
    \includegraphics[width=1\linewidth]{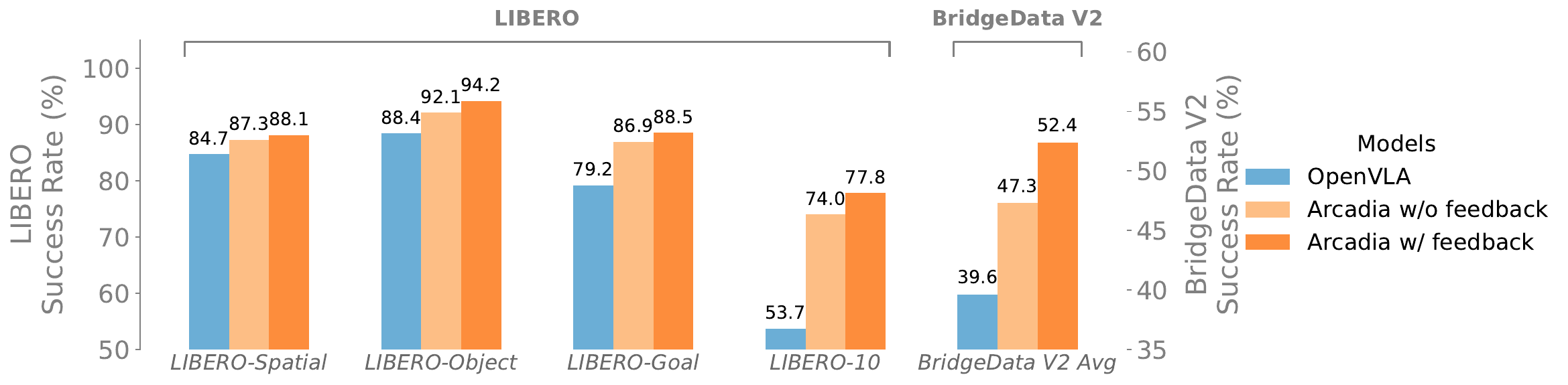}
    \vspace{-7mm}
    \caption{Comparison of various models and methods on different benchmarks.}
    \label{fig5}
    \vspace{-2mm}
\end{figure*}

\begin{figure*}
    \centering
    \includegraphics[width=1\linewidth]{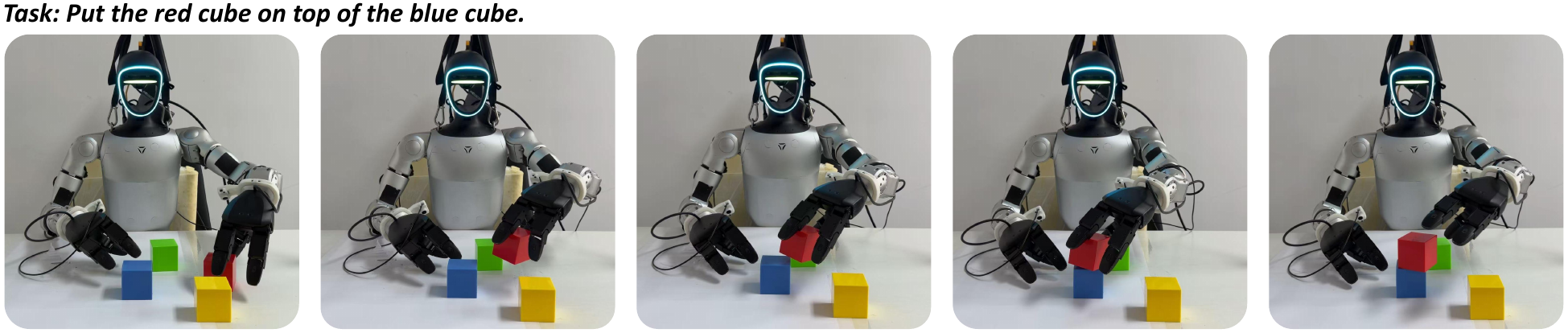}
    \vspace{-7mm}
    \caption{A representative case to illustrate Arcadia’s manipulation behavior.}
    \label{fig6}
    \vspace{-3mm}
\end{figure*}

\textbf{Backbone and Simulator.} \textbf{\texttt{Arcadia}} adopts the hierarchical architecture of NaVILA, composed of a high-level vision–language model (VLM) for perception and reasoning, and low-level controllers interfaced with robot APIs for execution in both simulation and the real world. We use \textit{Qwen2.5-VL}~\cite{Qwen2.5-VL} as the high-level VLM and the open-source Unitree G1~\cite{unitree_g1} platform for embodied validation. All simulated navigation data are organized in the VLN-CE~\cite{krantz2020beyond} format for cross-benchmark compatibility within Isaac Sim~\cite{nvidia_isaac_sim}.  

\textbf{Benchmarks and Evaluation.} We benchmark \textbf{\texttt{Arcadia}} on VLN-CE-Isaac~\cite{cheng2024navila}, R2R~\cite{anderson2018vision}, and RxR~\cite{ku2020room}, covering diverse instruction-following settings and visual domains. We also include ScanQA~\cite{azuma2022scanqa} for navigation-oriented question answering. Consolidated results are shown in Table~\ref{tab1}.

\textbf{Results.} Integrating QA-based reasoning into navigation enhances instruction comprehension, while \textbf{\texttt{Arcadia}}’s self-collected data provide greater environmental diversity and scene realism. Under identical architectures and training budgets, \textit{\textbf{\texttt{Arcadia}} w/o feedback} outperforms \textit{NaVILA} by an average success rate (SR) of 2.7\%, demonstrating the benefit of autonomous, task-aligned data generation. Incorporating real-world feedback (\textit{\textbf{\texttt{Arcadia}} w/ feedback}) further improves generalization and trajectory diversity, achieving the best performance across all benchmarks. Results in Table~\ref{tab1} confirm that these gains stem primarily from improved data quality and closed-loop refinement rather than dataset size alone.

\subsection{VLA Task Performance}
\label{sec4.2}

\textbf{Settings.} To assess \textbf{\texttt{Arcadia}}’s effectiveness on manipulation tasks, we compare three configurations:  
\begin{itemize}
\item \textit{OpenVLA}~\cite{kim2024openvla}. The baseline is trained in a single-stage setup using only the training dataset corresponding to the evaluation environment.
\item \textit{\textbf{\texttt{Arcadia}} w/o feedback.} Uses the same data scale as \textit{OpenVLA} but replaces all trajectories with manipulation data generated by \textbf{\texttt{Arcadia}}’s simulation pipeline.  
\item \textit{\textbf{\texttt{Arcadia}} w/ feedback.} The full closed-loop framework, where real-world feedback further refines policies and data, completing the real-to-sim-to-real cycle.  
\end{itemize}

\textbf{Backbone and Simulator.} \textbf{\texttt{Arcadia}} employs the same high-level vision–language model, \textit{Qwen2.5-VL}~\cite{Qwen2.5-VL}, augmented with an OpenVLA-style 7D action de-tokenizer. Simulated training is conducted in Isaac Sim~\cite{nvidia_isaac_sim}, and the GRUtopia~\cite{wang2024grutopia} robotic arm serves as the embodied platform for interaction and data collection within physically consistent scenes.

\textbf{Benchmarks and Evaluation.} We evaluate \textbf{\texttt{Arcadia}} across two major manipulation benchmarks: LIBERO~\cite{liu2023libero} and BridgeData (V2)~\cite{walke2023bridgedata}. To ensure compatibility, all datasets are standardized to the RLDS format. Evaluation metrics and scripts are aligned with their official implementations for consistency and fairness.  

\textbf{Results.} \textbf{\texttt{Arcadia}} outperforms \textit{OpenVLA} on all benchmarks. The improvements are particularly pronounced in BridgeData (V2), where feedback enhances object grounding and long-horizon stability. These results highlight \textbf{\texttt{Arcadia}}’s strong cross-task generalization and validate its scalability beyond language-conditioned navigation.

\subsection{Real-World Evaluation}
\label{sec4.3}

To evaluate \textbf{\texttt{Arcadia}}’s real-world performance, we established a physical testing environment for human evaluation, comprising 100 navigation and 100 manipulation tasks. Figures~\ref{fig4} and~\ref{fig6} illustrate representative configurations of this setup. It is noteworthy that all navigation tasks were conducted under a zero-shot setting, with no task-specific fine-tuning. In contrast, for manipulation tasks, we fine-tuned a dedicated dual-arm model to handle operations involving four blocks placed on a tabletop. Experiments were performed using a Unitree G1 robot equipped with a Dex-3 manipulator.

As baselines, they achieved 13 navigation and 9 manipulation successful completions, respectively, while \textbf{\texttt{Arcadia}} completed 46 and 27 tasks. Notably, in multi-destination navigation and multi-object manipulation scenarios, where both baselines failed, \textbf{\texttt{Arcadia}} maintained a 17\% success rate, often completing initial subtasks but struggling with extended or compositional instructions. These results indicate that although there is still room for improvement in physical deployment, \textbf{\texttt{Arcadia}} demonstrates markedly stronger generalization and reliability, underscoring its potential as a robust foundation for real-world embodied intelligence.

\subsection{Ablation and Analysis}
\label{sec4.4}

\textbf{Component Ablations.}  
In Table~\ref{tab2}, we conduct a detailed ablation study to analyze the contribution of each individual component within Arcadia. Specifically, we systematically replace one of its four key modules with a suboptimal alternative, allowing us to isolate and quantify the influence of each design choice on the overall system performance. The four components under examination are as follows:
(1) the static training set, in which the model is trained using the ScaleVLN and RLBench datasets;
(2) the retrieval-based scene reconstruction, where scenes are reconstructed using assets drawn from a limited asset library rather than a fully dynamic environment;
(3) the joint training strategy, in which the VLM backbone is optimized through multi-task learning to support both VLN and VLA tasks; and
(4) the sparse feedback mechanism, where the model receives only a binary success or failure signal as task feedback.

The results demonstrate that removing any single component significantly degrades overall performance, highlighting the importance of designing embodied intelligence systems that realistically capture the full lifecycle of interaction. Notably, the performance drop in case (3) is minimal, suggesting that the VLN and VLA tasks can share a unified VLM backbone. This finding motivates further exploration toward integrating VLN and VLA into a single framework, enabling unified execution of tasks that involve both navigation and manipulation.

\begin{figure}[!t]
    \centering
        \resizebox{0.95\linewidth}{!}{
        \renewcommand{\arraystretch}{1.2}
            \begin{tabular}{@{ }clcc@{ }}
            \midrule
            &  & \multicolumn{2}{c}{Success Rate (\%)} \\ 
            \cmidrule(lr){3-4}  
            &   &  VLN-CE-Isaac & LIBERO \\  
            \midrule
            0&Backbone & 44.9 & 76.5 \\
            \midrule
            1& Static training set & 43.0 & 72.9 \\
            2& Retrieval scene construction & 46.1 & 81.4\\
            3& Joint training & 49.8 & 87.0  \\
            4& Sparse feedback & 48.8 & 85.3 \\
            \midrule
            5& \textbf{\texttt{Arcadia}} & 50.1 & 87.2 \\
            \midrule
            \end{tabular}
        }
        \vspace{-3mm}
        \captionof{table}{Ablation study on each component.}
        \label{tab2}
        \vspace{-5mm}
\end{figure}

\section{Conclusion}

We introduced \textbf{\texttt{Arcadia}}, a full-lifecycle framework that unifies real-world data collection, generative simulation, shared representation learning, and feedback-based adaptation into a closed loop for embodied lifelong learning. By tightly coupling these stages, \textbf{\texttt{Arcadia}} enables robots to continuously refine both their environments and policies, achieving improvements across navigation and manipulation tasks in simulation and real-world deployments. The framework demonstrates strong generalization, robustness, and scalability, highlighting its potential as a practical foundation for developing self-improving embodied agents and advancing the study of real-to-sim-to-real learning.

\section{Limitations and Future Work}

While \textbf{\texttt{Arcadia}} provides a comprehensive full-lifecycle framework for embodied intelligence, the current implementation is confined to the Unitree G1 platform and the Isaac Sim simulator. Hardware availability and cost constraints further limit empirical validation to a 7B-scale vision–language model, thereby narrowing the scope of large-scale evaluation. Future work will extend \textbf{\texttt{Arcadia}} to a broader range of embodiments and simulation environments, incorporating emerging open frameworks such as InternRobot to strengthen the coupling between robots, tasks, and environments. In practice, a substantial portion of our development effort has been devoted to unifying interfaces and configuring diverse environments. Accordingly, our primary motivation is to enable researchers to focus on model innovation rather than engineering complexity. To this end, we maintain open-source releases to promote reproducibility and foster community collaboration, allowing researchers to concentrate on advancing foundational models instead of managing implementation overhead.

\clearpage
{
    \small
    \bibliographystyle{ieeenat_fullname}
    \bibliography{main}

@String(AAAI = {AAAI})

@article{baltes1987theoretical,
  title={Theoretical propositions of life-span developmental psychology: On the dynamics between growth and decline.},
  author={Baltes, Paul B},
  journal={Developmental psychology},
  volume={23},
  number={5},
  pages={611},
  year={1987},
  publisher={American Psychological Association}
}

@article{baltes1997incomplete,
  title={On the incomplete architecture of human ontogeny: Selection, optimization, and compensation as foundation of developmental theory.},
  author={Baltes, Paul B},
  journal={American psychologist},
  volume={52},
  number={4},
  pages={366},
  year={1997},
  publisher={American Psychological Association}
}

@book{kolb2014experiential,
  title={Experiential learning: Experience as the source of learning and development},
  author={Kolb, David A},
  year={2014},
  publisher={FT press}
}

@book{hebb2005organization,
  title={The organization of behavior: A neuropsychological theory},
  author={Hebb, Donald Olding},
  year={2005},
  publisher={Psychology press}
}

@article{lindenberger2019brain,
  title={Brain plasticity in human lifespan development: the exploration--selection--refinement model},
  author={Lindenberger, Ulman and L{\"o}vd{\'e}n, Martin},
  journal={Annual Review of Developmental Psychology},
  volume={1},
  number={1},
  pages={197--222},
  year={2019},
  publisher={Annual Reviews}
}

@article{parisi2019continual,
  title={Continual lifelong learning with neural networks: A review},
  author={Parisi, German I and Kemker, Ronald and Part, Jose L and Kanan, Christopher and Wermter, Stefan},
  journal={Neural networks},
  volume={113},
  pages={54--71},
  year={2019},
  publisher={Elsevier}
}

@book{chen2018lifelong,
  title={Lifelong machine learning},
  author={Chen, Zhiyuan and Liu, Bing},
  year={2018},
  publisher={Morgan \& Claypool Publishers}
}

@article{wilson2013embodied,
  title={Embodied cognition is not what you think it is},
  author={Wilson, Andrew D and Golonka, Sabrina},
  journal={Frontiers in psychology},
  volume={4},
  pages={58},
  year={2013},
  publisher={Frontiers Media SA}
}

@article{sullivan2018learning,
  title={Learning and embodied cognition: A review and proposal},
  author={Sullivan, Jaclynn V},
  journal={Psychology Learning \& Teaching},
  volume={17},
  number={2},
  pages={128--143},
  year={2018},
  publisher={Sage Publications Sage UK: London, England}
}

@book{pfeifer2006body,
  title={How the body shapes the way we think: a new view of intelligence},
  author={Pfeifer, Rolf and Bongard, Josh},
  year={2006},
  publisher={MIT press}
}

@article{xiao2025robot,
  title={Robot learning in the era of foundation models: A survey},
  author={Xiao, Xuan and Liu, Jiahang and Wang, Zhipeng and Zhou, Yanmin and Qi, Yong and Jiang, Shuo and He, Bin and Cheng, Qian},
  journal={Neurocomputing},
  pages={129963},
  year={2025},
  publisher={Elsevier}
}

@article{mon2025embodied,
  title={Embodied large language models enable robots to complete complex tasks in unpredictable environments},
  author={Mon-Williams, Ruaridh and Li, Gen and Long, Ran and Du, Wenqian and Lucas, Christopher G},
  journal={Nature Machine Intelligence},
  pages={1--10},
  year={2025},
  publisher={Nature Publishing Group UK London}
}

@inproceedings{zitkovich2023rt,
  title={Rt-2: Vision-language-action models transfer web knowledge to robotic control},
  author={Zitkovich, Brianna and Yu, Tianhe and Xu, Sichun and Xu, Peng and Xiao, Ted and Xia, Fei and Wu, Jialin and Wohlhart, Paul and Welker, Stefan and Wahid, Ayzaan and others},
  booktitle={Conference on Robot Learning},
  pages={2165--2183},
  year={2023},
  organization={PMLR}
}

@inproceedings{o2024open,
  title={Open x-embodiment: Robotic learning datasets and rt-x models: Open x-embodiment collaboration 0},
  author={O’Neill, Abby and Rehman, Abdul and Maddukuri, Abhiram and Gupta, Abhishek and Padalkar, Abhishek and Lee, Abraham and Pooley, Acorn and Gupta, Agrim and Mandlekar, Ajay and Jain, Ajinkya and others},
  booktitle={2024 IEEE International Conference on Robotics and Automation (ICRA)},
  pages={6892--6903},
  year={2024},
  organization={IEEE}
}

@article{team2024octo,
  title={Octo: An open-source generalist robot policy},
  author={Team, Octo Model and Ghosh, Dibya and Walke, Homer and Pertsch, Karl and Black, Kevin and Mees, Oier and Dasari, Sudeep and Hejna, Joey and Kreiman, Tobias and Xu, Charles and others},
  journal={arXiv preprint arXiv:2405.12213},
  year={2024}
}

@inproceedings{bharadhwaj2024roboagent,
  title={Roboagent: Generalization and efficiency in robot manipulation via semantic augmentations and action chunking},
  author={Bharadhwaj, Homanga and Vakil, Jay and Sharma, Mohit and Gupta, Abhinav and Tulsiani, Shubham and Kumar, Vikash},
  booktitle={2024 IEEE International Conference on Robotics and Automation (ICRA)},
  pages={4788--4795},
  year={2024},
  organization={IEEE}
}

@article{bousmalis2023robocat,
  title={Robocat: A self-improving foundation agent for robotic manipulation},
  author={Bousmalis, Konstantinos and Vezzani, Giulia and Rao, Dushyant and Devin, Coline and Lee, Alex X and Bauza, Maria and Davchev, Todor and Zhou, Yuxiang and Gupta, Agrim and Raju, Akhil and others},
  journal={arXiv preprint arXiv:2306.11706},
  volume={1},
  number={8},
  year={2023}
}

@article{wang2024grutopia,
  title={Grutopia: Dream general robots in a city at scale},
  author={Wang, Hanqing and Chen, Jiahe and Huang, Wensi and Ben, Qingwei and Wang, Tai and Mi, Boyu and Huang, Tao and Zhao, Siheng and Chen, Yilun and Yang, Sizhe and others},
  journal={arXiv preprint arXiv:2407.10943},
  year={2024}
}

@article{cheng2024navila,
  title={Navila: Legged robot vision-language-action model for navigation},
  author={Cheng, An-Chieh and Ji, Yandong and Yang, Zhaojing and Gongye, Zaitian and Zou, Xueyan and Kautz, Jan and B{\i}y{\i}k, Erdem and Yin, Hongxu and Liu, Sifei and Wang, Xiaolong},
  journal={arXiv preprint arXiv:2412.04453},
  year={2024}
}

@article{chang2017matterport3d,
  title={Matterport3d: Learning from rgb-d data in indoor environments},
  author={Chang, Angel and Dai, Angela and Funkhouser, Thomas and Halber, Maciej and Niessner, Matthias and Savva, Manolis and Song, Shuran and Zeng, Andy and Zhang, Yinda},
  journal={arXiv preprint arXiv:1709.06158},
  year={2017}
}

@inproceedings{savva2019habitat,
  title={Habitat: A platform for embodied ai research},
  author={Savva, Manolis and Kadian, Abhishek and Maksymets, Oleksandr and Zhao, Yili and Wijmans, Erik and Jain, Bhavana and Straub, Julian and Liu, Jia and Koltun, Vladlen and Malik, Jitendra and others},
  booktitle={Proceedings of the IEEE/CVF international conference on computer vision},
  pages={9339--9347},
  year={2019}
}

@online{nvidia_isaac_sim,
  author  = {{NVIDIA Corporation}},
  title   = {NVIDIA Isaac Sim},
  year    = {2025},
  url     = {https://developer.nvidia.com/isaac/sim},
  urldate = {2025-10-30},
  note    = {Official product page}
}

@online{nvidia_isaac_lab,
  author  = {{NVIDIA Corporation}},
  title   = {NVIDIA Isaac Lab},
  year    = {2025},
  url     = {https://developer.nvidia.com/isaac/lab},
  urldate = {2025-10-30},
  note    = {Official product page}
}

@article{jaafar2024lanmp,
  title={Lanmp: A language-conditioned mobile manipulation benchmark for autonomous robots},
  author={Jaafar, Ahmed and Raman, Shreyas Sundara and Wei, Yichen and Juliani, Sofia and Wernerfelt, Anneke and Quartey, Benedict and Idrees, Ifrah and Liu, Jason Xinyu and Tellex, Stefanie},
  journal={arXiv preprint arXiv:2412.05313},
  year={2024}
}

@inproceedings{shridhar2020alfred,
  title={Alfred: A benchmark for interpreting grounded instructions for everyday tasks},
  author={Shridhar, Mohit and Thomason, Jesse and Gordon, Daniel and Bisk, Yonatan and Han, Winson and Mottaghi, Roozbeh and Zettlemoyer, Luke and Fox, Dieter},
  booktitle={Proceedings of the IEEE/CVF conference on computer vision and pattern recognition},
  pages={10740--10749},
  year={2020}
}

@inproceedings{padmakumar2022teach,
  title={Teach: Task-driven embodied agents that chat},
  author={Padmakumar, Aishwarya and Thomason, Jesse and Shrivastava, Ayush and Lange, Patrick and Narayan-Chen, Anjali and Gella, Spandana and Piramuthu, Robinson and Tur, Gokhan and Hakkani-Tur, Dilek},
  booktitle={Proceedings of the AAAI Conference on Artificial Intelligence},
  volume={36},
  number={2},
  pages={2017--2025},
  year={2022}
}

@inproceedings{anderson2018vision,
  title={Vision-and-language navigation: Interpreting visually-grounded navigation instructions in real environments},
  author={Anderson, Peter and Wu, Qi and Teney, Damien and Bruce, Jake and Johnson, Mark and S{\"u}nderhauf, Niko and Reid, Ian and Gould, Stephen and Van Den Hengel, Anton},
  booktitle={Proceedings of the IEEE conference on computer vision and pattern recognition},
  pages={3674--3683},
  year={2018}
}

@article{ku2020room,
  title={Room-across-room: Multilingual vision-and-language navigation with dense spatiotemporal grounding},
  author={Ku, Alexander and Anderson, Peter and Patel, Roma and Ie, Eugene and Baldridge, Jason},
  journal={arXiv preprint arXiv:2010.07954},
  year={2020}
}

@inproceedings{shridhar2022cliport,
  title={Cliport: What and where pathways for robotic manipulation},
  author={Shridhar, Mohit and Manuelli, Lucas and Fox, Dieter},
  booktitle={Conference on robot learning},
  pages={894--906},
  year={2022},
  organization={PMLR}
}

@article{james2020rlbench,
  title={Rlbench: The robot learning benchmark \& learning environment},
  author={James, Stephen and Ma, Zicong and Arrojo, David Rovick and Davison, Andrew J},
  journal={IEEE Robotics and Automation Letters},
  volume={5},
  number={2},
  pages={3019--3026},
  year={2020},
  publisher={IEEE}
}

@inproceedings{ross2011reduction,
  title={A reduction of imitation learning and structured prediction to no-regret online learning},
  author={Ross, St{\'e}phane and Gordon, Geoffrey and Bagnell, Drew},
  booktitle={Proceedings of the fourteenth international conference on artificial intelligence and statistics},
  pages={627--635},
  year={2011},
  organization={JMLR Workshop and Conference Proceedings}
}

@article{meng2025preserving,
  title={Preserving and combining knowledge in robotic lifelong reinforcement learning},
  author={Meng, Yuan and Bing, Zhenshan and Yao, Xiangtong and Chen, Kejia and Huang, Kai and Gao, Yang and Sun, Fuchun and Knoll, Alois},
  journal={Nature Machine Intelligence},
  pages={1--14},
  year={2025},
  publisher={Nature Publishing Group UK London}
}

@inproceedings{jakobi1995noise,
  title={Noise and the reality gap: The use of simulation in evolutionary robotics},
  author={Jakobi, Nick and Husbands, Phil and Harvey, Inman},
  booktitle={European conference on artificial life},
  pages={704--720},
  year={1995},
  organization={Springer}
}

@inproceedings{tobin2017domain,
  title={Domain randomization for transferring deep neural networks from simulation to the real world},
  author={Tobin, Josh and Fong, Rachel and Ray, Alex and Schneider, Jonas and Zaremba, Wojciech and Abbeel, Pieter},
  booktitle={2017 IEEE/RSJ international conference on intelligent robots and systems (IROS)},
  pages={23--30},
  year={2017},
  organization={IEEE}
}

@inproceedings{peng2018sim,
  title={Sim-to-real transfer of robotic control with dynamics randomization},
  author={Peng, Xue Bin and Andrychowicz, Marcin and Zaremba, Wojciech and Abbeel, Pieter},
  booktitle={2018 IEEE international conference on robotics and automation (ICRA)},
  pages={3803--3810},
  year={2018},
  organization={IEEE}
}

@article{andrychowicz2020learning,
  title={Learning dexterous in-hand manipulation},
  author={Andrychowicz, OpenAI: Marcin and Baker, Bowen and Chociej, Maciek and Jozefowicz, Rafal and McGrew, Bob and Pachocki, Jakub and Petron, Arthur and Plappert, Matthias and Powell, Glenn and Ray, Alex and others},
  journal={The International Journal of Robotics Research},
  volume={39},
  number={1},
  pages={3--20},
  year={2020},
  publisher={SAGE Publications Sage UK: London, England}
}

@article{de2021continual,
  title={A continual learning survey: Defying forgetting in classification tasks},
  author={De Lange, Matthias and Aljundi, Rahaf and Masana, Marc and Parisot, Sarah and Jia, Xu and Leonardis, Ale{\v{s}} and Slabaugh, Gregory and Tuytelaars, Tinne},
  journal={IEEE transactions on pattern analysis and machine intelligence},
  volume={44},
  number={7},
  pages={3366--3385},
  year={2021},
  publisher={IEEE}
}

@article{khetarpal2022towards,
  title={Towards continual reinforcement learning: A review and perspectives},
  author={Khetarpal, Khimya and Riemer, Matthew and Rish, Irina and Precup, Doina},
  journal={Journal of Artificial Intelligence Research},
  volume={75},
  pages={1401--1476},
  year={2022}
}

@misc{isaac_ros,
  title        = {NVIDIA Isaac ROS: A collection of GPU-accelerated ROS2 packages for autonomous robots},
  howpublished = {\url{https://github.com/NVIDIA-ISAAC-ROS}},
  year         = {2025},
  author       = {{NVIDIA Corporation}},
}

@misc{isaac_ros_nvblox,
  title        = {Isaac ROS Nvblox: GPU-Accelerated 3D Reconstruction for ROS 2},
  author       = {{NVIDIA Corporation}},
  howpublished = {\url{https://github.com/NVIDIA-ISAAC-ROS/isaac_ros_nvblox}},
  year         = {2025}
}

@article{mao2025spatiallm,
  title={SpatialLM: Training Large Language Models for Structured Indoor Modeling},
  author={Mao, Yongsen and Zhong, Junhao and Fang, Chuan and Zheng, Jia and Tang, Rui and Zhu, Hao and Tan, Ping and Zhou, Zihan},
  journal={arXiv preprint arXiv:2506.07491},
  year={2025}
}

@article{xu2024instantmesh,
  title={Instantmesh: Efficient 3d mesh generation from a single image with sparse-view large reconstruction models},
  author={Xu, Jiale and Cheng, Weihao and Gao, Yiming and Wang, Xintao and Gao, Shenghua and Shan, Ying},
  journal={arXiv preprint arXiv:2404.07191},
  year={2024}
}

@inproceedings{krantz2020beyond,
  title={Beyond the nav-graph: Vision-and-language navigation in continuous environments},
  author={Krantz, Jacob and Wijmans, Erik and Majumdar, Arjun and Batra, Dhruv and Lee, Stefan},
  booktitle={European Conference on Computer Vision},
  pages={104--120},
  year={2020},
  organization={Springer}
}

@inproceedings{walke2023bridgedata,
  title={Bridgedata v2: A dataset for robot learning at scale},
  author={Walke, Homer Rich and Black, Kevin and Zhao, Tony Z and Vuong, Quan and Zheng, Chongyi and Hansen-Estruch, Philippe and He, Andre Wang and Myers, Vivek and Kim, Moo Jin and Du, Max and others},
  booktitle={Conference on Robot Learning},
  pages={1723--1736},
  year={2023},
  organization={PMLR}
}

@article{Qwen2.5-VL,
  title={Qwen2.5-VL Technical Report},
  author={Bai, Shuai and Chen, Keqin and Liu, Xuejing and Wang, Jialin and Ge, Wenbin and Song, Sibo and Dang, Kai and Wang, Peng and Wang, Shijie and Tang, Jun and Zhong, Humen and Zhu, Yuanzhi and Yang, Mingkun and Li, Zhaohai and Wan, Jianqiang and Wang, Pengfei and Ding, Wei and Fu, Zheren and Xu, Yiheng and Ye, Jiabo and Zhang, Xi and Xie, Tianbao and Cheng, Zesen and Zhang, Hang and Yang, Zhibo and Xu, Haiyang and Lin, Junyang},
  journal={arXiv preprint arXiv:2502.13923},
  year={2025}
}

@misc{unitree_g1,
  title        = {Unitree G1 Humanoid Robot},
  author       = {{Unitree Robotics}},
  year         = {2025},
  url          = {https://www.unitree.com/g1},
}

@inproceedings{azuma2022scanqa,
  title={Scanqa: 3d question answering for spatial scene understanding},
  author={Azuma, Daichi and Miyanishi, Taiki and Kurita, Shuhei and Kawanabe, Motoaki},
  booktitle={proceedings of the IEEE/CVF conference on computer vision and pattern recognition},
  pages={19129--19139},
  year={2022}
}

@article{liu2023libero,
  title={Libero: Benchmarking knowledge transfer for lifelong robot learning},
  author={Liu, Bo and Zhu, Yifeng and Gao, Chongkai and Feng, Yihao and Liu, Qiang and Zhu, Yuke and Stone, Peter},
  journal={Advances in Neural Information Processing Systems},
  volume={36},
  pages={44776--44791},
  year={2023}
}

@article{kim2024openvla,
  title={Openvla: An open-source vision-language-action model},
  author={Kim, Moo Jin and Pertsch, Karl and Karamcheti, Siddharth and Xiao, Ted and Balakrishna, Ashwin and Nair, Suraj and Rafailov, Rafael and Foster, Ethan and Lam, Grace and Sanketi, Pannag and others},
  journal={arXiv preprint arXiv:2406.09246},
  year={2024}
}
}

\end{document}